\documentclass[twocolumn,10pt,a4paper]{article}

\usepackage[margin=0.75in, columnsep=0.25in]{geometry} 
\usepackage{times}
\usepackage[utf8]{inputenc}
\usepackage{hyperref}
\usepackage{booktabs}
\usepackage{microtype}

\usepackage{algorithm}
\usepackage{algpseudocode}
\usepackage{mathtools}
\usepackage{amssymb}
\usepackage{tikz}
\usepackage{tabularx}
\usepackage{enumitem}
\usepackage[most]{tcolorbox}
\usepackage{xcolor}
\usepackage{multirow}

\usepackage[
  style=numeric,
  sorting=none,
  backend=biber
  ]{biblatex}

\usepackage{microtype}

\usetikzlibrary{positioning}
\usetikzlibrary{arrows.meta}
  
\newtcolorbox{takeawaybox}{
  colback=black!3,
  colframe=black!25,
  boxrule=0.4pt,
  arc=2pt,
  left=6pt, right=6pt, top=4pt, bottom=4pt,
  width=\columnwidth,
  fonttitle=\bfseries,
  title=Key Takeaways,
}

\setlist[enumerate]{parsep=0pt,listparindent=\parindent}
\newcolumntype{L}{>{\raggedright\arraybackslash}X}

\begin{document}

\twocolumn[
  \begin{@twocolumnfalse}
    \begin{center}
        \vspace{0.4cm}
        {\Large \textbf{Towards Trustworthy Wi-Fi CSI-based Sensing: \\ Systematic Evaluation of Adversarial Robustness}} \\
        \vspace{0.5cm}

        \textbf{Shreevanth Krishnaa Gopalakrishnan\footnotemark} \quad and \quad \textbf{Stephen Hailes} \\
        \vspace{0.1cm}
        Department of Computer Science \\
        University College London \\
        \texttt{\{shreevanth.gopalakrishnan, s.hailes\}@ucl.ac.uk}
    \end{center}
    
    \vspace{0.3cm}

    \begin{center}
    \begin{minipage}{0.9\textwidth} 
        \hrule height 1pt 
        \vspace{0.2cm}
        \textbf{Abstract} \\
        Machine learning drives Channel State Information (CSI)-based human sensing in modern wireless networks, enabling applications like device-free human activity recognition (HAR) and identification (HID). However, the susceptibility of these models to adversarial perturbations raises security concerns that must be quantified prior to edge deployment. 

We present a systematic robustness evaluation of five diverse CSI architectures across four public datasets, jointly analyzing white-box, black-box transfer, and universal attacks, together with defense strategies, under unconstrained and physics-guided perturbation boundaries. Contrary to prior assumptions, our experiments reveal that model capacity does not guarantee robustness; simple architectures consistently exhibit superior resilience compared to high-capacity sequence and vision models. Furthermore, vulnerability is fundamentally task-dependent, with HAR proving highly susceptible to attack, while HID demonstrates stark inherent resistance. Crucially, enforcing physical signal constraints drastically reduces attack success rates and significantly taxes attacker computation, showing that standard unconstrained feature-space attacks substantially overestimate real-world Over-The-Air vulnerabilities. By synthesizing attack, defense, and security metrics with strict edge hardware considerations, this work establishes foundational design principles for secure, deployable, and physically realizable wireless sensing systems.
        \vspace{0.3cm}
        
        \vspace{0.3cm}
        \hrule height 0.5pt 
    \end{minipage}
    \end{center}
    
    \vspace{0.3cm} 
  \end{@twocolumnfalse}
]
\footnotetext{Corresponding author}

\section{Introduction}
With the imminent standardization of IEEE 802.11bf and the evolution of 6G, wireless sensing is transitioning from a supplementary network feature into ubiquitous infrastructure. As the demand for intelligent environments grows---spanning homes, offices, airports, and care facilities---the number of use cases depending on accurate human positioning, detection, and recognition has increased dramatically. Channel State Information (CSI)-based sensing has proven particularly promising in this domain, enabling device-free, fine-grained, and partially privacy-preserving sensing using commodity Wi-Fi devices. By capturing multipath channel characteristics, CSI provides the high-resolution perception necessary for human activity recognition (HAR), localization, and identification.

Early Wi-Fi sensing was pioneered by the Intel 5300 CSI Tool \cite{halperin_tool_2011}, which exposed per-subcarrier data but required modified drivers and offline processing. Recently, platforms such as Atheros \cite{xie_precise_2019}, Nexmon \cite{nexmon:project}, ESP32 \cite{hernandez_lightweight_2020}, and ZTECSITool \cite{wang_wi-fi_2025} have democratized real-time, low-cost sensing without specialized hardware. However, as these systems are increasingly deployed in security-sensitive environments, a holistic analysis of their ability to withstand malicious over-the-air (OTA) interference in the physical (PHY) and digital domains is lacking.

\subsection{Key Research Issues}

Machine Learning (ML) is central to mapping their temporal and spectral channel variations to human-centric tasks. Yet, these models are inherently vulnerable to adversarial perturbations. While extensively studied in computer vision (CV) \cite{goodfellow_explaining_2015}, PHY attacks are fundamentally different: wireless perturbations must survive a physical medium governed by multipath fading, spectral masking, and phase continuity. These channel dynamics make unconstrained digital noise---common in CV research---highly unrealistic in a wireless context. Though existing work has explored attacks at the PHY and network layers (Figure \ref{fig:threat-taxonomy}), the ultimate goal remains the exploitation of fragile ML decision boundaries.

Despite the growing number of CSI datasets, a systematic evaluation of model robustness is still lacking. Prior works such as WiAdv \cite{zhou_wiadv_2022}, WiIntruder \cite{cao_security_2024}, and SecureSense \cite{yang_securesense_2024} demonstrate that perturbations can degrade HAR performance, but they vary significantly in experimental setup, model scale, and attack assumptions. Moreover, few studies incorporate physics-inspired constraints, such as channel correlation and Doppler coherence, which influences whether a perturbation is actually realizable. This motivates the need for a \textbf{unified robustness evaluation framework} that adapts general ML adversarial techniques to the unique physical, mathematical and Radio Frequency (RF) signal-processing requirements of CSI-based sensing.

\subsection{Comparison with the State of the Art}

Several surveys and technical papers trace the evolution of wireless sensing from Received Signal Strength Indicator (RSSI) to CSI, highlighting its societal potential in future Wi-Fi and cellular networks. However, the majority of these works focus primarily on performance under benign conditions \cite{ahmad_wifi-based_2024, zheng_zero-effort_2019, zhang_device-free_2021, tan_commodity_2022, nirmal_deep_2021, ma_wifi_2020, geng_survey_2024, chen_cross-domain_2023}. Complementary studies aim to improve raw accuracy and generalization through refined processing pipelines, environment-independent representations, or cross-domain adaptation \cite{xin_freesense_2018, zhang_wi-run_2018, meneghello_sharp_2023, wang_rt-fall_2017, wang_phasebeat_2017, ma_signfi_2018, guo_wiar_2019}. More recently, robustness-oriented research has explored the effects of signal injections, universal perturbations, and adversarial training \cite{xu_wicam_2022, yang_securesense_2024, zhou_wiadv_2022, sharma_wi-spoof_2025, yin_evasion_2025, zhou_adversarial_2020}. These efforts are typically confined to specific models or attack families and lack standardized evaluations. Furthermore, the development of algorithmic defenses to harden CSI ML models remains a secondary priority compared to exposing vulnerabilities.

\subsection{Summary of Novel Contributions}

To fill these critical gaps, this work provides the first systematic evaluation of machine learning adversarial robustness for CSI-based human sensing under realistic wireless constraints and diverse threat models. To guide the design of secure and trustworthy Wi-Fi sensing systems, we:

\begin{enumerate}
    \item \textbf{Formulate Physics-Aware Threat Models:} We introduce physically constrained attacks that preserve spectral, spatial, and temporal correlations, establishing a realistic Over-The-Air (OTA) baseline.
    \item \textbf{Conduct a Comprehensive Benchmark:} We evaluate white-box, black-box, and universal attacks across diverse CSI datasets and architectures for various sensing tasks such as human activity recognition and human identification.
    \item \textbf{Evaluate Mitigation Feasibility:} We analyze standard defenses (e.g., PGD-AT and Randomized Smoothing) against RF-specific attacks, exposing the trade-offs between empirical robustness, clean accuracy, and hardware latency for Edge/TinyML platforms.
    \item \textbf{Release an Open-Source Artifact:} We provide a modular evaluation framework to the community to enable reproducible research across new datasets, models, attack vectors and defenses. 
\end{enumerate}

\section{Background}
This section provides an overview of how Wi-Fi signals can be repurposed for device-free human-centric sensing. It further introduces the emerging security and privacy challenges, culminating in recent efforts to characterize and defend against adversarial attacks on time-series models.

\subsection{WLAN-based sensing}
\label{subsec: wlan_based_sensing}

Wireless Local Area Network (WLAN)-based sensing has gained significant attention as it enables localization and activity recognition using existing Wi-Fi infrastructure. Three main approaches exist: Received Signal Strength Indicator (RSSI), Passive Wi-Fi Radar (PWR), and CSI-based sensing~\cite{li_csi_2022}. All are based on the principle that environmental dynamics alter wireless propagation through multipath effects, frequency shifts and attenuation.  

RSSI is coarse-grained, susceptible to multipath fading even in static environments, and challenging to use in robust detectors. PWR, inspired by bistatic radar, extracts parameters such as Angle of Arrival (AoA), Time of Flight (ToF), Time Difference of Arrival (TDoA), and Doppler Frequency Shift (DFS). Despite algorithmic maturity, PWR faces physical constraints: a 2.4\,GHz Wi-Fi channel with 20\,MHz bandwidth offers insufficient ToF resolution for fine-grained indoor sensing; self-interference also remains a limitation~\cite{yang_rssi_2013}. Finally, while CSI offers the highest fidelity by capturing subcarrier-level transformations, its sensitivity to environmental non-stationarity and hardware impairments necessitates rigorous phase and amplitude calibration \cite{ma_wifi_2020}.

The IEEE 802.11 family defines Wi-Fi standards for WLAN technologies~\cite{du_overview_2023}. While Wi-Fi sensing has been explored for over a decade, earlier standards did not explicitly support it. The recent IEEE 802.11bf amendment \cite{noauthor_ieee_2025} introduces a landmark change: Wi-Fi is now formalized as a joint communication and sensing technology. 802.11bf specifies new PHY and MAC layer mechanisms to enable standardized sensing measurements, facilitating interoperability across devices. 

\subsection{Channel State Information (CSI)}

\begin{figure*}[t]
    \centering
    \includegraphics[width=0.75\linewidth]{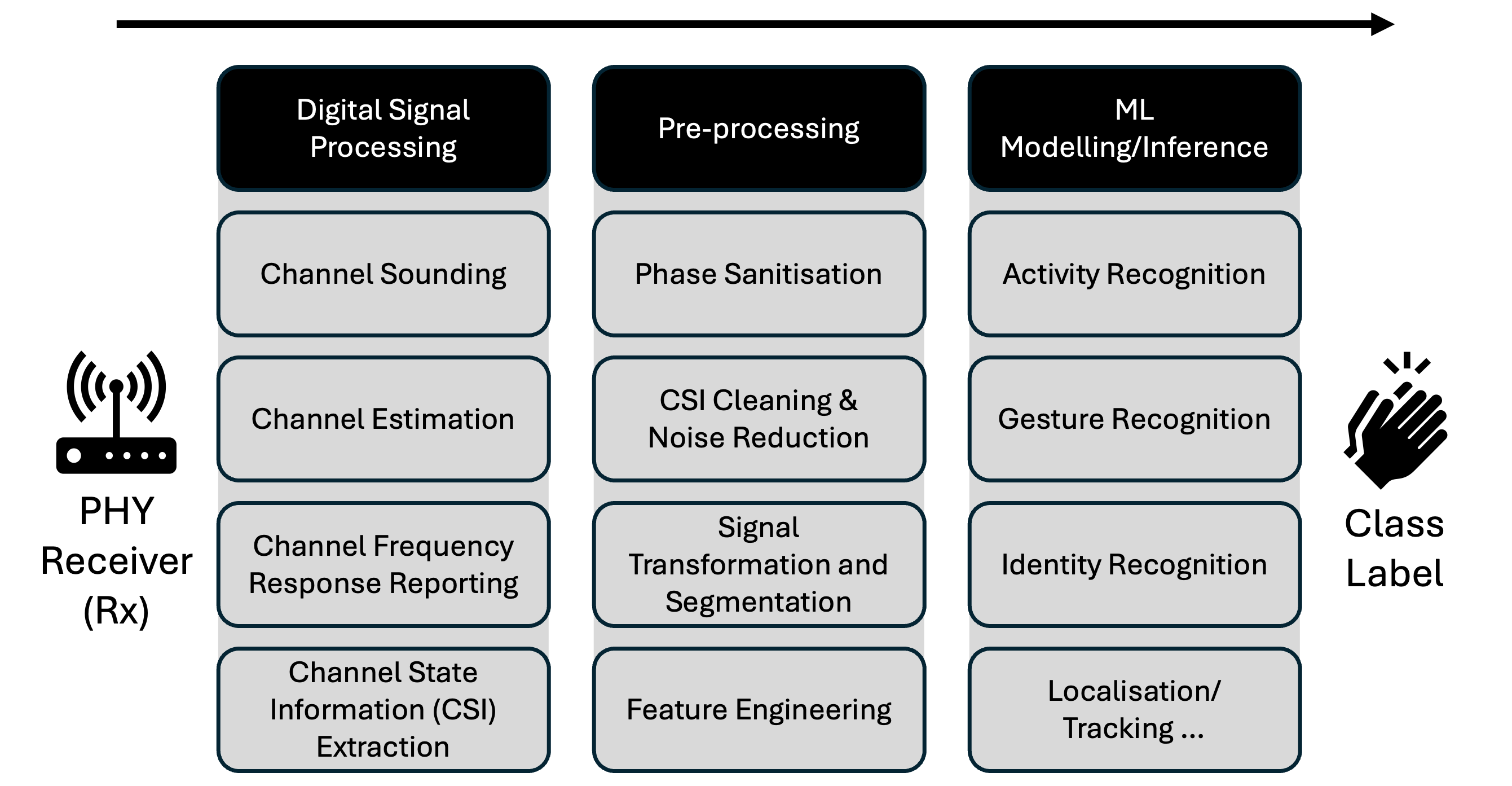}
    \caption{CSI processing pipeline---from signal reception to classification result}
    \label{fig:csi_pipeline}
\end{figure*}

In Orthogonal Frequency-Division Multiplexing (OFDM)-based Wi-Fi systems, the channel between transmitter (Tx) and receiver (Rx) can be represented as a 3-D CSI tensor:

\begin{equation}
    \mathbf{H} \in \mathbb{C}^{N_r \times N_t \times K}
\end{equation}

where $N_r$ and $N_t$ denote the number of receive and transmit antennas, and $K$ is the number of subcarriers. Each scalar entry is a complex coefficient capturing amplitude attenuation $|H_{i,j}(k)|$ and phase shift $\angle H_{i,j}(k)$ for subcarrier $k$ on the $(i,j)$ antenna link:

\begin{equation}
    H_{i,j}(k) = |H_{i,j}(k)| , e^{\mathrm{j} \angle H_{i,j}(k)}
\end{equation}

CSI is typically sampled at high rates (1–10\,kHz), forming time-series that encode variations in the propagation channel. Human presence, gestures, and activities leave distinctive temporal and spatial signatures in the CSI amplitude and phase. Since these patterns are not directly interpretable, machine learning methods are employed for tasks such as presence detection, fall detection, and activity recognition.  

The end-to-end CSI processing pipeline, visualized in Figure \ref{fig:csi_pipeline}, begins at the \emph{physical (PHY) layer}, where transmitted Wi-Fi packets are received by one or more antennas and the channel is estimated using pilot symbols embedded in each OFDM frame. The receiver computes the Channel Frequency Response (CFR) across all subcarriers and antennas, yielding complex CSI coefficients that capture amplitude attenuation and phase shift caused by multipath propagation. These raw CSI measurements are then passed through a sequence of \emph{pre-processing} operations such as phase sanitization, denoising, filtering, and normalization to remove hardware-induced offsets and noise. Depending on the application, additional \emph{signal transformations} (e.g., short-time Fourier transform, Doppler or body-velocity profiles~\cite{zheng_zero-effort_2019}) may be applied to derive motion-sensitive representations. The resulting feature tensors form the input to a \emph{machine learning model}, typically a convolutional or a recurrent network, which maps temporal and/or spatial channel variations to semantic outputs such as human activity, gesture, or identity labels. The \emph{classification result} is produced at the application layer and can support downstream tasks such as authentication and occupancy detection \cite{ma_wifi_2020}. 

\subsection{Threat Landscape in Wi-Fi Sensing}

\begin{figure}[t]
\centering
\begin{tikzpicture}[font=\scriptsize, node distance=2mm]
  \tikzstyle{layer}=[draw, rounded corners=1pt, line width=0.3pt, align=left, inner sep=2pt, text width=0.9\columnwidth]
  \node[layer] (L1) {\textbf{1) Physical layer:} signal injection, jamming, replay, hardware tampering};
  \node[layer, above=of L1] (L2) {\textbf{2) Network layer:} packet modification, dropping, timing manipulation};
  \node[layer, above=of L2] (L3) {\textbf{3) Data / Pre-processing:} phase sanitization exploits, transform-domain perturbations, segmentation or window manipulation};
  \node[layer, above=of L3] (L4) {\textbf{4) ML model:} model extraction, poisoning, backdoors, \textbf{evasion}};
  \node[layer, above=of L4] (L5) {\textbf{5) System / Deployment:} privacy leakage, cross-device transfer, confidence manipulation};
\end{tikzpicture}
\caption{Threat model taxonomy for CSI-based sensing}
\label{fig:threat-taxonomy}
\end{figure}

The Wi-Fi sensing pipeline introduces multiple points of vulnerability, spanning the wireless protocol stack and the ML hierarchy (Figure \ref{fig:threat-taxonomy}). At the \emph{physical and network layers}, adversaries can deploy signal injection, jamming, or packet-level manipulation to corrupt CSI acquisition. Moving upward, the \emph{pre-processing layer} is susceptible to phase-sanitization exploits and transform-domain perturbations, while the \emph{ML model layer}---the primary focus of our work---remains inherently vulnerable to training-time poisoning and inference-time evasion. Finally, at the \emph{system layer}, exploits such as cross-device transfer or privacy leakage can compromise end-to-end trustworthiness. The imminent standardization of IEEE 802.11bf makes understanding these multi-layer risks---from gesture-based authentication spoofing to the disruption of safety-critical fall detection of elderly people---increasingly urgent. 

The same fidelity that enables rich activity recognition also amplifies ethical risks of surveillance and adversarial exploitation. Adopting the perspective of Geng et al. \cite{geng_survey_2024}, Wi-Fi sensing can be viewed through three interconnected lenses. As a \emph{Sword}, CSI can be weaponized as a side-channel to infer identity and presence without consent or perturbed to induce malicious misclassifications. As a \emph{Shield}, it necessitates the development of robust models and filtering strategies to harden sensing against such interference. Ultimately, the user or system can also be a \emph{Victim}, facing potential privacy loss through covert occupancy detection, security breaches via authentication spoofing, or safety hazards when adversarial noise disrupts critical HAR alarms.

\subsection{Adversarial Machine Learning}

Within Adversarial Machine Learning (AML), the category of attacks where an adversary manipulates test-time inputs to induce misclassification is termed \textit{evasion}. According to a recent NIST report on AML, evasion attacks are broadly classified by the threat model into \textit{white-box} and \textit{black-box}~\cite{vassilev_adversarial_2025}.  

In the white-box setting, the attacker has full knowledge of the target model (architecture, hyperparameters, and training data). This represents a strong attacker and is the most commonly studied threat model, as it allows evaluation under worst-case assumptions. Conversely, the black-box setting assumes only minimal access, e.g., query interfaces, representing a more realistic but weaker adversarial position. We can taxonomize them further as follows~\cite{pitropakis_taxonomy_2019}: \\

\textit{\textbf{White-box evasion attacks:}}
\begin{itemize}
    \item \textbf{Optimization-based attacks}, where adversaries solve a constrained problem to find a bounded perturbation that induces misclassification:
    \begin{flalign}
        && \mathbf{x}^{\ast} = \mathbf{x} + \boldsymbol{\delta} \quad \text{s.t.} \ \|\boldsymbol{\delta}\|_p \le \epsilon, \ \arg\max_k [f_\theta(\mathbf{x}^{\ast})]_k \neq y
    \end{flalign}
    where $\mathbf{x}$ is the input, $f_\theta$ the model, $y$ the ground-truth, and $\epsilon$ the perturbation budget for the $\ell_p$ norm.
    
    \item \textbf{Universal perturbations}, refer to a single perturbation vector transferable across most inputs, typically via iterative optimization on a representative subset.  
    
    \item \textbf{Physically realizable attacks}, designed to survive domain distortions. In wireless sensing, a difficulty is that perturbations injected in the RF domain may not map linearly to the model input space due to quantization, filtering, and multipath distortion.
\end{itemize}

\textit{\textbf{Black-box evasion attacks:}}  
\begin{itemize}
    \item \textbf{Score-based attacks}, where adversaries use model logits or probabilities to approximate gradients.  
    \item \textbf{Decision-based attacks}, where adversaries access only the final predicted labels and iteratively approximate decision boundaries.  
    \item \textbf{Transfer-based attacks}, where adversaries train surrogate models or ensembles and transfer adversarial examples to the target system.  
\end{itemize}

In response, several defense strategies have been proposed: \textbf{Adversarial Training} \cite{madry_towards_2018} hardens decision boundaries using generated attacks; \textbf{Regularization-based Methods} \cite{tack_consistency_2021} enforce prediction consistency; \textbf{Input Transformations} \cite{guo_countering_2018, zhang_countering_2021} attempt to denoise inputs before inference; and \textbf{Certified Defenses} \cite{wong_provable_2018, cohen_certified_2019} offer provable guarantees within $\ell_p$ perturbation bounds. However, directly porting these to wireless sensing often fails due to unique RF channel dynamics and computational limits of Edge hardware.

\section{Related Work}
\subsection{Overview}

While early work prioritized sensing fidelity, accuracy and scalability, recent efforts emphasize advanced modeling pipelines---environment generalization~\cite{wang_transfer_2023, meneghello_sharp_2023, li__2024}, few-/zero-shot learning~\cite{yin_fewsense_2022, wang_review_2024} and cross-application frameworks \cite{zhao_crossfi_2025, wang_airfi_2024}---and edge and/or federated deployments for heterogeneous IoT contexts~\cite{wen_integrated_2025, jeon_compressive_2021, ramadan_federated_2025}. Real-world systems are now expected to be adaptable to new users, environments, and configurations without exhaustive retraining. 

In parallel, a growing body of work has been exploring how these systems behave when exposed to malicious interference or perturbations. However, security and privacy research in this area has been fragmented---spanning PHY layer manipulation, crafted RF signal injections, and adversarial ML attacks. To clarify this landscape, we begin by synthesizing prior works into \emph{attack classes}.

\subsection{Attack Classes for Wireless Sensing}
\label{subsec: attack_classes_hcws}

Several studies have proposed attack vectors and highlighted the vulnerabilities of Wi-Fi-based sensing systems to adversaries aiming to undermine privacy, integrity, or availability. Adapting prior taxonomies~\cite{yang_securesense_2024}, ~\cite{sun_sok_2024}, these threats can be broadly grouped as follows: \textbf{inference attacks}, \textbf{signal injection attacks} and \textbf{digital/adversarial perturbations}.

\subsubsection{\textbf{Inference Attacks}}
Attackers can infer private information such as human presence, gestures, or even physiological signals without consent, exploiting the very properties that make Wi-Fi sensing convenient: open standards, ubiquity, device-free/passive operation, and minimal user cooperation. These attacks range in scale from:
\begin{itemize}
    \item \textit{Small:} keystroke detection ~\cite{li_when_2016} and vital sign monitoring, e.g., heart rate, respiration, sleep position~\cite{duan_comprehensive_2023}.
    \item \textit{Medium:} gesture/activity recognition \cite{gu_wigrunt_2022, zheng_zero-effort_2019, zhu_et_2020}.
    \item \textit{Large:} people counting \cite{ropitault_overhead-free_2024}, occupancy detection \cite{yang_device-free_2018}, gait-based/person identification \cite{wang_wipin_2019, xin_freesense_2016}.
\end{itemize}

Attackers with transmission capabilities can escalate from passive inference to active disruption by injecting crafted signals into the environment. These signal injections are, by definition, \emph{physically realizable}, since they operate directly in the wireless channel and therefore must obey its propagation, synchronization, and power constraints. By distorting the measured channel state, such attacks can degrade or deny the availability of legitimate sensing systems \cite{qiao_phycloak_2016, sharma_wi-spoof_2025}. Similar mechanisms may also be used defensively, e.g., injecting controlled noise to mask sensitive user motion or suppress unwanted inference \cite{argyriou_obfuscation_2023}.

A range of OTA attack and defense mechanisms demonstrate how adversaries can steer receiver-side CSI while preserving normal communication links. Examples include Aegis, a privacy-preserving RF shielding system \cite{yao_aegis_2018}; FooLoc, which models multiplicative propagation effects under channel constraints for localization and fingerprinting \cite{xiao_over--air_2023}; WiAdv, which performs gesture spoofing using full-duplex radios \cite{zhou_wiadv_2022}; and Li et al.'s work \cite{li_practical_2024}, in which they modify Wi-Fi preambles to corrupt channel estimation. Collectively, these works highlight the need for physics-guided constraints---such as coherence time/bandwidth, spatial correlation, and PSR limits---when defining realistic threat models.

\subsubsection{\textbf{Adversarial Perturbations in the Signal Domain}}
Adversarial Machine Learning extends these ideas to the data and model layers. Early gradient-based attacks remain the most widely used \cite{goodfellow_explaining_2015, madry_towards_2018, carlini_towards_2017}, and although designed for vision tasks, many have since been adapted to sequential data~\cite{zizzo_adversarial_2020, pialla_time_2025}, where perturbations must respect temporal smoothness and signal continuity. Applying such threats to Wi-Fi sensing introduces additional challenges: CSI is complex-valued, high-dimensional, and non-stationary. Prior work shows that both white- and black-box attacks can reliably degrade model performance \cite{yin_evasion_2025, ambalkar_adversarial_2021}, yet most methods ignore physical-layer constraints such as channel propagation, multipath structure, and transceiver behavior, limiting real-world feasibility. A smaller subset does account for these factors---e.g., WiCAM’s surrogate-based black-box attacks \cite{xu_wicam_2022}, and WiIntruder’s universal perturbations which are resilient to synchronization offsets \cite{cao_security_2024}. Despite these recent advances, the field still lacks a unified robustness evaluation framework for offense and defense.

\subsubsection{\textbf{Assumptions and Threat Models}}
Most works assume an attacker can place an SDR or Wi-Fi transceiver in the environment. Depending on capability, the adversary may be \textit{Rx-only} (passive) or \textit{Tx/Rx} (active OTA injection), and may have \textit{white-box} access to models/data or operate in \textit{black-box/transfer} regimes. Our evaluation adopts this layered perspective to assess robustness across diverse attacker capabilities.

\subsection{Lightweight Modeling for Wi-Fi Sensing}
\label{subsec: edge_sensing_rw}

According to the emerging IEEE 802.11bf standard, CSI will be captured and processed directly on access points and IoT devices. As edge devices typically operate under tight latency, memory, and energy constraints, and large-scale environments may require dozens of sensing nodes, this greatly increases the attack surface. These realities make lightweight, efficient and robust models essential for real-time operation, multi-node scalability, and privacy-preserving on-device processing. Youm and Go~\cite{youm_lightweight_2025} adopt neural architecture search and structured pruning to design compact HAR models; LiteHAR~\cite{salehinejad_litehar_2022} replaces heavy backpropagation with random convolution kernels and ridge regression; and LiteWiHAR~\cite{liu_litewihar_2024} targets embedded devices with a small-footprint design.

Yet, existing approaches primarily focus on architectural compression and inference-time efficiency. Much less attention has been given to the deeper modeling challenge inherent to Wi-Fi sensing: CSI provides extremely high-dimensional, complex-valued inputs, yet datasets remain relatively small and difficult to collect. Lightweight models that lack robustness may fail unpredictably under distribution shifts or malicious perturbations, underscoring the need to consider efficiency and robustness together.

\subsection{Summary and Motivation}
Existing research has explored the effects of signal injections and adversarial attacks. However, these studies often isolate specific models or datasets and neglect the physical realism of perturbations. This paper builds upon these foundations by establishing a unified framework for CSI-based sensing, incorporating physics-inspired constraints and multiple attack regimes to provide a reproducible evaluation of adversarial robustness in human-centric Wi-Fi sensing.

\section{Evaluation Framework}
\begin{figure}
    \centering
    \includegraphics[width=\linewidth]{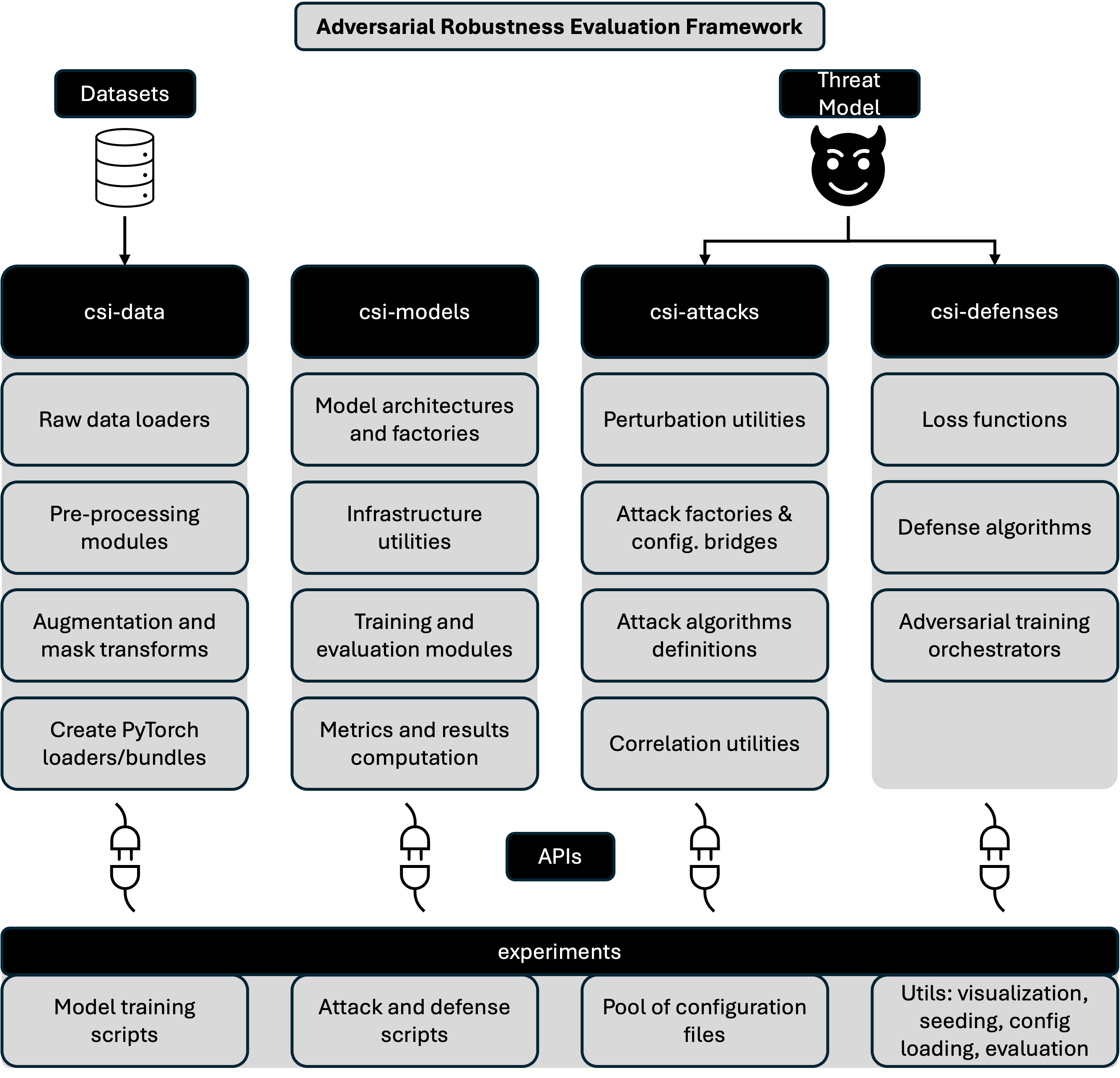}
    \caption{Modular framework under the \texttt{csi} namespace}
    \label{fig:exp_software_stack}
\end{figure}

Our \emph{Adversarial Robustness Evaluation Framework}, shown in Figure \ref{fig:exp_software_stack}, is organized into modular layers that mirror the end-to-end Wi-Fi sensing and CSI ML adversarial robustness workflow. \texttt{csi-data} handles dataset ingestion, preprocessing, and augmentation through configurable loader bundles. \texttt{csi-models} defines the learning backbone, providing model factories, training loops, and evaluation metrics. \texttt{csi-attacks} encapsulates gradient- and score-based adversarial algorithms, along with correlation-preserving and physically grounded perturbation utilities. \texttt{csi-defenses} implements countermeasures, providing shared loss modules and adversarial trainers. These components form the foundation for the datasets, tools and algorithms, providing the infrastructure for our experiments.

The APIs are unified under an \texttt{experiments} layer that manages configuration files, training and evaluation scripts, and cluster orchestration for large-scale studies. Together, the framework enables reproducible experimentation on CSI datasets, supporting clean and adversarial evaluations, performance profiling, and systematic analysis of robustness under varying threat models. The entire framework is implemented in \texttt{Python}, and relies on libraries such as \texttt{PyTorch}, \texttt{NumPy}, \texttt{Matplotlib}, and \texttt{YAML}. An open-source release will be made available upon acceptance.

\section{Methodology}
This paper improves upon SenseFi \cite{yang_sensefi_2023}, a recent benchmark which we augmented by adding ML models along with modules for adversarial attack and defense. 

\subsection{Threat Model}

Our threat model focuses on the edge ML devices where CSI data is processed. In practice, an adversary can compromise these systems via two primary vectors: injecting malicious OTA RF signals or exploiting the edge hardware itself (e.g., via OS-level malware or hardware Trojans). Because the receiver's PHY processing---including Automatic Gain Control (AGC) and synchronization---is highly non-differentiable, calculating a worst-case OTA attack via classical optimization is mathematically intractable. We therefore model the impact of either vector directly in the digital domain: specifically, after CSI extraction but before ML inference. This approach establishes a strict mathematical upper bound for adversarial disruption. To attempt to bridge the gap between this theoretical bound and physical reality, we enforce constraints on the digital perturbations as a mathematical proxy for the wireless channel and receiver filters, ensuring the resulting attacks resemble realizable interference.

\begin{figure*}[t]
    \centering
    \includegraphics[width=0.75\linewidth]{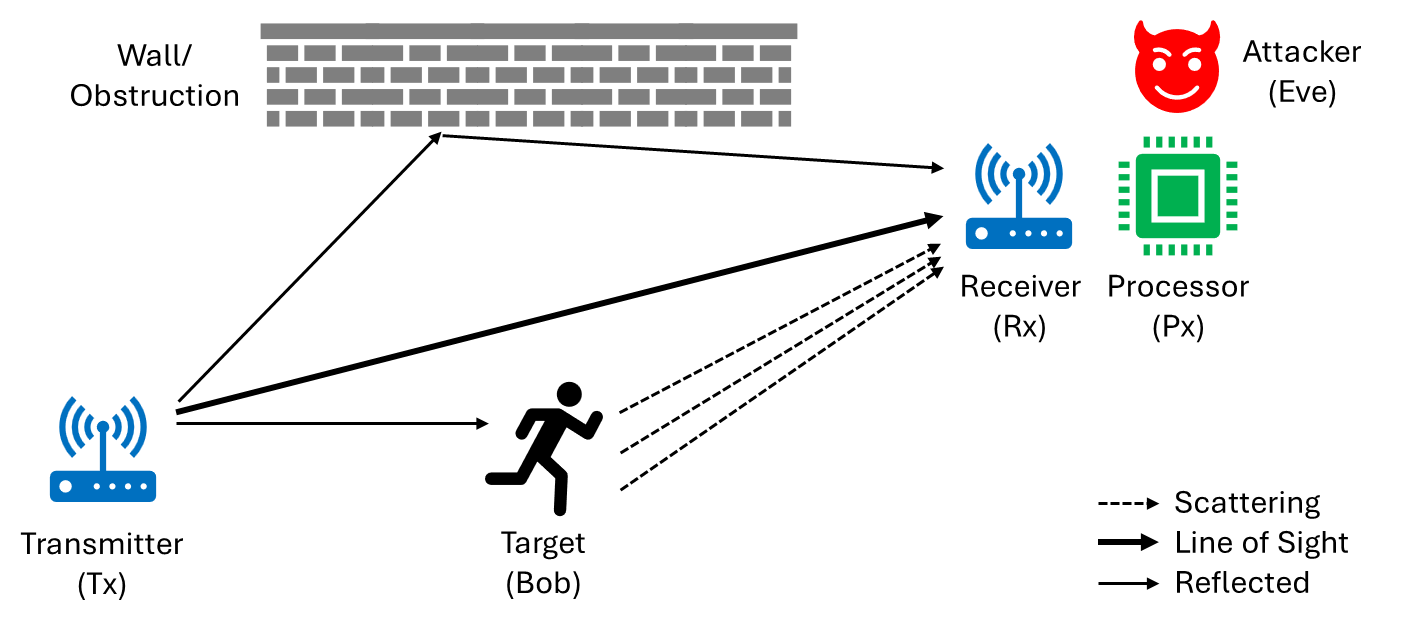}
    \caption{Threat model. Px is compromised by Eve}
    \label{fig:threat_model}
\end{figure*}

As illustrated in Figure \ref{fig:threat_model}, the system comprises a transmitter (\emph{Tx}), a receiver (\emph{Rx}), a sensing target (\emph{Bob}), and an adversary (\emph{Eve}). We assume Eve is co-located with the edge platform (\emph{Px}) and seeks to induce misclassification while remaining stealthy. Stealth is achieved by crafting perturbations that satisfy fundamental physical constraints of real Wi-Fi signals, thereby mimicking legitimate transmissions. As our framework evaluates algorithmic robustness directly on the digital CSI arrays, the attack generation remains agnostic to the physical geometry of the sensing environment.

\subsection{Datasets}

Table \ref{tab:datasets} summarizes the datasets used for our robustness benchmarks: NTU-Fi (HAR \& HID) \cite{yang_efficientfi_2022, wang_caution_2022} and CSI-Bench (HAR \& HID) \cite{zhu_csi-bench_2025}. We selected these to provide a rigorous contrast between controlled laboratory environments and \emph{in-the-wild} deployment scenarios. The NTU-Fi datasets are established, high-resolution standards captured using the Atheros CSI tool, offering a reliable baseline. In contrast, CSI-Bench introduces significant complexity through both platform and device heterogeneity. Originally collected across 26 diverse real-world environments using 16 different device types, CSI-Bench forces models to contend with the natural signal variability and hardware impairments typical of actual residential and office deployments. Each dataset was partitioned into training, validation, and testing sets in a 7:2:1 ratio. The raw CSI amplitudes were $z$-score normalized on a per-subcarrier basis to mitigate hardware-specific power variations. Furthermore, the NTU-Fi datasets underwent temporal smoothing to reduce high-frequency noise. Batch sizes varied from 32--128 (training--testing) for NTU-Fi and were set to 256 for CSI-Bench.

\begin{table*}[t]
\centering
\caption{CSI-based datasets for Wi-Fi sensing}
\label{tab:datasets}
\setlength{\tabcolsep}{0pt} 
\begin{tabularx}{\linewidth}{@{\extracolsep{\fill}} c *{4}{>{\centering\arraybackslash}X} @{}}
\toprule
\textbf{Datasets} & \textbf{NTU-Fi-HAR} & \textbf{NTU-Fi-HID} & \textbf{CSI-Bench-HAR} & \textbf{CSI-Bench-HID} \\
\midrule
Environment & \multicolumn{2}{c}{Controlled Lab} & \multicolumn{2}{c}{In-the-wild (5 Envs.)} \\
Platform & \multicolumn{2}{c}{Atheros CSI Tool} & \multicolumn{2}{c}{Commercial IoT (e.g., Apple HomePod, Google Nest)} \\
Hardware Diversity & \multicolumn{2}{c}{Homogeneous (NIC)} & \multicolumn{2}{c}{Heterogeneous (e.g., Wyze, Govee, Eightree Plugs)} \\
Data Format (C, F, T) & \multicolumn{2}{c}{(3, 114, 250)} & \multicolumn{2}{c}{(1, 56, 250)} \\
\midrule
Label Count & 6 Activities & 14 Subjects & 5 Activities & 5 Subjects \\
Total Samples & 1,200 & 840 & 22,636 & 11,485 \\
\bottomrule
\end{tabularx}
\end{table*}

\subsection{Models}

To systematically evaluate adversarial robustness across diverse structural assumptions, we implemented five supervised architectures: \textbf{CSIMLP} (MLP), a simple baseline that processes flattened channel-frequency-time (C, F, T) tensors through custom residual blocks; \textbf{CSILeNet} (LeNet), a lightweight 2D CNN for local spatial-temporal feature extraction; \textbf{CSIResNet-18} (ResNet18), a deep hierarchical network adapted to accept variable CSI input channels; \textbf{CSIBiLSTM} (BiLSTM), a sequence model that captures temporal dependencies across per-timestep features; and \textbf{CSITimeSformer} (TimeSFormer), a modern architecture applying divided temporal and spatial attention to 2D patches. 

For compatibility across heterogeneous datasets, we developed custom model factories that automatically configure input dimensions, normalization layers, and classification heads. Each model was trained for up to 80 epochs using the Adam optimizer (learning rate $10^{-3}$, weight decay $5\times10^{-4}$) with early stopping based on validation loss. 

\subsection{Adversarial Attacks}

We evaluated a suite of adversaries that represent strong and practical threats against sensing infrastructure. Concretely, we tested: (i) first-order \textbf{white-box attacks} to probe gradient sensitivity across architectures and datasets, (ii) \textbf{transfer attacks} to quantify black-box risk from model-stealing or surrogate-training scenarios, and (iii) \textbf{universal adversarial perturbations (UAPs)}~\cite{moosavi-dezfooli_universal_2017} to evaluate input-agnostic interference patterns across samples. To align with realistic threat models, our evaluation is task-specific: HAR is evaluated under an \textit{untargeted} setting to simulate general availability attacks, while HID is evaluated under a \textit{targeted} setting to simulate integrity attacks such as biometric spoofing. All attacks are evaluated in \textit{constrained} and \textit{unconstrained} forms.

\subsubsection{\textbf{Formulation}}
\label{subsubsec: attack_formulation}

Let $\mathcal{D} = \{(\mathbf{x}_i, y_i)\}_{i=1}^N$ be a labeled CSI dataset comprising $N$ samples, where $\mathbf{x}_i \in \mathcal{X}$ denotes a multidimensional CSI vector or matrix, and $y_i \in \{1, \dots, K\}$ denotes its discrete ground-truth label among $K$ distinct classes. A classifier $f_\theta: \mathcal{X} \to \mathbb{R}^K$ produces a $K$-dimensional vector of logits $f_\theta(\mathbf{x})$, with the final discrete prediction being $\hat{y} = \arg\max_k [f_\theta(\mathbf{x})]_k$. 

Traditionally, an \textit{additive} adversarial perturbation $\boldsymbol{\delta}$ alters the input via $\mathbf{x}_{adv} = \mathbf{x} + \boldsymbol{\delta}$, bounded by an $\ell_p$ norm $\|\boldsymbol{\delta}\|_p \le \varepsilon$. In some wireless contexts~\cite{li_practical_2024}, physical-layer RF vulnerabilities (e.g., pilot contamination) induce \textit{multiplicative} perturbations, formulated as $\mathbf{x}_{adv} = \alpha \mathbf{x}$ and constrained by hardware limits $\alpha \in [\alpha_{\min}, \alpha_{\max}]$. We employ the $\ell_p$-bounded additive formulation as a rigorous mathematical superset---since any multiplicative perturbation $\alpha$ can be expressed as $\boldsymbol{\delta} = (\alpha - 1)\mathbf{x}$---and optimize this objective using APGD, the gold standard for robustness evaluation~\cite{croce_reliable_2020}. Depending on the threat model, the adversary either seeks to simply induce misclassification (\textit{untargeted}, $\hat{y}(\mathbf{x}_{adv}) \neq y$) or strictly force the model to output a predetermined class (\textit{targeted}, $\hat{y}(\mathbf{x}_{adv}) = y_{\text{target}}$). We refer the reader to the original papers for full algorithmic details.

\subsubsection{\textbf{Auto-Projected Gradient Descent (APGD)}}

APGD is a ``parameter-free'' attack that extends vanilla PGD by eliminating the need for manual step-size tuning. Instead, it introduces a trend-aware adaptive step size $\eta^{(t)}$ and a momentum parameter $\beta$. Letting $\Pi_{\mathcal{S}}$ denote the projection onto the feasible set, the update step can be defined as:

\begin{equation}
    \begin{aligned}
        \mathbf{z}^{(t+1)} &= \Pi_{\mathcal{S}} \big( \boldsymbol{\delta}^{(t)} + \eta^{(t)} \cdot \mathbf{u}^{(t)} \big) \\
        \boldsymbol{\delta}^{(t+1)} &= \Pi_{\mathcal{S}} \Bigl( \boldsymbol{\delta}^{(t)} + \beta \big(\mathbf{z}^{(t+1)} - \boldsymbol{\delta}^{(t)}\big) \\
        &\quad + (1-\beta)\big(\boldsymbol{\delta}^{(t)} - \boldsymbol{\delta}^{(t-1)}\big) \Bigr)
    \end{aligned}
\end{equation}
        
Here, $\mathbf{u}^{(t)} = \nabla_{\boldsymbol{\delta}} \mathcal{L} / \|\nabla_{\boldsymbol{\delta}} \mathcal{L}\|_2$ is the $\ell_2$-normalized gradient direction of the chosen objective function $\mathcal{L}$. The step size $\eta^{(t)}$ is halved at specific checkpoints if the objective fails to improve sufficiently over recent iterations. The algorithm tracks and returns the perturbation yielding the highest objective value across all iterations and restarts. In our context, the projection $\Pi_{\mathcal{S}}$ encapsulates not only the relative energy budget but also the physical channel constraints, formulated explicitly as $\Pi_{\mathrm{phys}}$ in the subsequent sections.

\subsubsection{\textbf{Targeted APGD (APGD-T)}}
For targeted evasion, the algorithm utilizes the same APGD update steps, but the gradient direction $\mathbf{u}^{(t)}$ is derived from a targeted objective $\mathcal{L}_{\mathrm{target}}$, which drives the model toward a specific target class $y_{\mathrm{target}} \neq y$:
\begin{equation}
    \boldsymbol{\delta}^{\star} = \arg\max_{\boldsymbol{\delta} \in \mathcal{S}} \mathcal{L}_{\mathrm{target}} \big( f_{\theta}(\mathbf{x}+\boldsymbol{\delta}),\, y,\, y_{\mathrm{target}} \big)
\end{equation}
When the target sweep is enabled, the attack evaluates the top-$K$ non-true candidate classes derived from the clean logits, aggregating successes across targets and restarts. Otherwise, it executes fixed-target optimization.

\subsubsection{\textbf{Universal Adversarial Perturbations (UAP)}}
Rather than on a per-sample basis, we optimize a single, input-agnostic perturbation $\mathbf{v}$ directly via mini-batch gradient ascent. To ensure the perturbation stays physically constrained, it is bounded by the dataset's mean signal norm $\bar{r}$:
\begin{equation}
    \mathbf{v}^{(t+1)} = \Pi_{\|\cdot\|_2 \le \epsilon_2 \bar{r}} \Bigg( \mathbf{v}^{(t)} + \gamma \frac{\nabla_{\mathbf{v}} J(\mathbf{v}^{(t)};\mathcal{B})} {\|\nabla_{\mathbf{v}} J(\mathbf{v}^{(t)};\mathcal{B})\|_2} \Bigg)
\end{equation}

Here, $J(\mathbf{v}; \mathcal{B}) = \mathbb{E}_{\mathbf{x} \in \mathcal{B}} [\mathcal{L}(\mathbf{x} + \mathbf{v})]$ represents the batch-averaged objective function. $\mathcal{B}$ denotes a mini-batch of clean samples used to estimate the gradient, $\gamma$ is the optimization step size, and $\epsilon_2$ is the relative budget derived from the target Signal-to-Noise Ratio (SNR).

\subsubsection{\textbf{Perturbation Constraints}}
\label{para: pert_constraints}
We adopt the Euclidean ($\ell_2$) norm to measure the distance between clean ($\mathbf{x}$) and adversarial ($\boldsymbol{\delta}$) signals, as it maps directly to signal energy. Thus, we define the Perturbation-to-Signal Ratio (PSR) as:

\begin{equation}
    \mathrm{PSR} = -\mathrm{SNR} = 10 \log_{10}\left(\frac{\|\boldsymbol{\delta}\|_2^2}{\|\mathbf{x}\|_2^2}\right)
\end{equation}

Higher PSR values indicate stronger perturbations relative to the SNR. In practice, an attack budget specified as a target SNR is converted into an absolute per-sample norm bound $\varepsilon$:
\begin{equation}
    \varepsilon = \varepsilon_{\mathrm{rel}}\|\mathbf{x}\|_2, \quad \varepsilon_{\mathrm{rel}} = 10^{-\mathrm{SNR}_{\mathrm{dB}}/20}
\end{equation}

\subsubsection{\textbf{Physically Constrained Projection Operator}}
To remain stealthy and physically realizable, adversarial RF perturbations must respect the indoor Wi-Fi channel's physical constraints, as unrealistic deviations in CSI would be easily detectable. Thus, we instantiate the abstract projection $\Pi_{\mathcal{S}}$ below as a composite physics-guided projection operator, detailed in Algorithm \ref{alg:physproj}, which explicitly maps mathematical projections to underlying physical phenomena such as multipath spreading, coherence time, and MIMO correlation.

\begin{equation}
    \Pi_{\mathrm{phys}}(\boldsymbol{\delta}) = \Pi_{\ell_2}\Big(\Pi_{\text{spatial}} \circ \Pi_{\text{temporal}} \circ \Pi_{\text{frequency}}(\boldsymbol{\delta})\Big)
\end{equation}

\begin{algorithm}[H]
\caption{Physically Constrained Projection $\Pi_{\mathrm{phys}}(\boldsymbol{\delta})$}
\label{alg:physproj}
\begin{algorithmic}[1]
\Require Perturbation $\boldsymbol{\delta} \in \mathbb{R}^{K\times T\times N_a}$, clean CSI $\mathbf{X}$, bound $\varepsilon$
\Ensure Projected, physically plausible perturbation $\tilde{\boldsymbol{\delta}}$

\Statex \textbf{Frequency Correlation (PDP-based)}
\Statex \textit{Spreads energy across subcarriers consistent with coherence bandwidth. Under WSSUS assumption, frequency correlation function is Fourier transform of Power Delay Profile (PDP) \cite{tse_fundamentals_2005}.}
\State Construct Toeplitz matrix $C$ matching the PDP:
\begin{equation*}
    C_{ij} = \begin{cases} 
    \exp\!\big(-(2\pi\tau_{\mathrm{rms}}\lvert f_i-f_j\rvert)^2\big), & \text{Gaussian} \\
    \big(1+(2\pi\tau_{\mathrm{rms}}\lvert f_i-f_j\rvert)^2\big)^{-1/2}, & \text{Exponential} 
    \end{cases}
\end{equation*}
\State Normalize $\mathrm{diag}(C)=1$, then apply $\boldsymbol{\delta} \leftarrow C\boldsymbol{\delta}$.

\Statex \textbf{Temporal Smoothing (Doppler Constraint)}
\Statex \textit{Captures how slowly time-varying channels evolve across OFDM symbols (coherence time $\propto$ 1/Doppler).}
\State Apply 1D convolution along time $t$: 
\begin{equation*}
    \boldsymbol{\delta}(k,t) \leftarrow (\boldsymbol{\delta}(k,\cdot) * g_{\sigma_t})(t), \quad g_{\sigma_t}(t) \propto e^{-t^2/(2\sigma_t^2)}
\end{equation*}

\Statex \textbf{Spatial/MIMO Correlation}
\Statex \textit{Models overlapping propagation paths across antenna elements via receive covariance matrix \cite{wang_spatial-temporal_2007}.}
\State Estimate receive covariance $\mathbf{R}_{\mathrm{rx}}=\mathbb{E}[\mathbf{h}\mathbf{h}^{\mathrm{H}}]$ from $\mathbf{X}$.
\State Factorize $\mathbf{R}_{\mathrm{rx}}=\mathbf{L}\mathbf{L}^{\mathrm{H}}$ and apply per $(k,t)$: $\boldsymbol{\delta}_{k,t,:} \leftarrow \boldsymbol{\delta}_{k,t,:}\,\mathbf{L}^{\mathrm{T}}$.

\Statex \textbf{Distributional Alignment (MMD Penalty)}
\Statex \textit{Prevents manifold divergence using RBF-kernel MMD, keeping perturbed samples close to clean data manifold \cite{gretton_kernel_2012, chen_dynamic_2022, zhao_knn-mmd_2024}.}
\State Penalize divergence between $\mathbf{X}$ and $\mathbf{Y}=\mathbf{X}+\boldsymbol{\delta}$:
\begin{align*}
    \widehat{\mathrm{MMD}}^2(\mathbf{X},\mathbf{Y}) &= \frac{1}{m(m-1)}\sum_{i\neq j}\big[k(\mathbf{x}_i,\mathbf{x}_j)+k(\mathbf{y}_i,\mathbf{y}_j)\big] \\
    &\quad - \frac{2}{m^2}\sum_{i,j}k(\mathbf{x}_i,\mathbf{y}_j)
\end{align*}

\Statex \textbf{$\ell_2$ Normalization and Projection}
\Statex \textit{Strictly bounds total energy to specified PSR, consistent with OTA practice \cite{li_practical_2024, xiao_over--air_2023}.}
\begin{equation*}
    \tilde{\boldsymbol{\delta}} = \varepsilon \boldsymbol{\delta} / \max(\varepsilon, \|\boldsymbol{\delta}\|_2)
\end{equation*}
\State \Return $\tilde{\boldsymbol{\delta}}$
\end{algorithmic}
\end{algorithm}

\subsubsection{\textbf{Implementation Details}}
White-box and transfer attacks were generated with budgets of 5, 10, 20, and 40\,dB SNR, but UAPs were evaluated only at 10\,dB. White-box attacks ran for 100 steps with 5 random restarts using random initialization. For these, HAR utilized untargeted APGD with cross-entropy loss, while HID employed targeted APGD-T with Difference of Logits Ratio loss and a top-5 target sweep. Transfer attacks followed a similar configuration but were limited to 30 steps, with HID utilizing label-wise APGD. UAPs were APGD-based, utilizing 8 outer passes with $\gamma = 0.1$, a target fooling rate of 0.9, and 50 mini-batches to estimate $\bar{r}$. In constrained settings, white-box and transfer attacks incorporated an MMD term with a weight of $w_{\mathrm{MMD}} = 0.05$.

\subsection{Adversarial Defenses}
\label{sec:defenses}

We evaluated two diverse robustness augmentation techniques: \textbf{Adversarial Training} and \textbf{Certified Defense}. We use the same notation as in \S\ref{subsubsec: attack_formulation}.

\subsubsection{\textbf{Adversarial Training (APGD-AT)}}

The standard adversarial training objective is formulated as a min-max optimization problem \cite{madry_towards_2018}:
\begin{equation}
    \min_\theta\;\mathbb{E}_{(\mathbf{x},y)\sim\mathcal{D}}\Big[\max_{\|\boldsymbol{\delta}\|_2\le\varepsilon\|\mathbf{x}\|_2}\mathcal{L}\big(f_\theta(\mathbf{x}+\boldsymbol{\delta}),y\big)\Big].
\end{equation}

The inner maximization finds the worst-case perturbations (with APGD), while the outer minimization updates the model parameters $\theta$ on those generated adversarial examples.

\subsubsection{\textbf{Randomized Smoothing}}

To construct a certifiably robust classifier $g$ from the base network $f_\theta$ \cite{cohen_certified_2019}, we uniquely inject domain-scaled isotropic Gaussian noise $\boldsymbol{\eta} \sim \mathcal{N}(0, \frac{\sigma_d^2}{2} \mathbf{I})$ into the inputs during training. At test time, $g(\mathbf{x})$ returns the expected prediction over this noise distribution:

\begin{equation}
    g(\mathbf{x}) = \arg\max_c \mathbb{P}_{\boldsymbol{\eta}}\big(f_\theta(\max(\mathbf{x} + \boldsymbol{\eta}, 0)) = c\big).
\end{equation}

This guarantees that the prediction remains constant within an $\ell_2$ radius around $\mathbf{x}$, determined by the noise variance $\sigma_d^2$ and the probability margin of the top class.

\subsubsection{\textbf{Implementation Details}}
APGD-AT fine-tuned clean, pretrained models under a 20\,dB SNR perturbation budget using 100 APGD inner steps, 3 restarts, and random initialization. Randomized smoothing was trained at the same 20\,dB operating point using amplitude noise with a domain-specific $\sigma$, two noisy replicas per sample, and a clean-loss mixing factor of 0.1. Both defenses were optimized using Adam (learning rate $10^{-3}$, weight decay $0$) for 20 epochs.

\section{Results}
Owing to the large number of configurations/parameters to sweep, each setup was executed on distributed cluster compute nodes, each with up to 32GB of RAM and high-performance GPUs (e.g., NVIDIA L40S, RTX6000 Ada, etc.). 

\subsection{Clean Model Baseline}
\label{sec:clean-eval}
We evaluated the clean performance of all models across the two HAR datasets (NTU-Fi-HAR and CSI-Bench-HAR) and the two HID datasets (NTU-Fi-HID and CSI-Bench-HID). Table~\ref{tab:clean_csi_ntu_supervised} summarizes the macro F1-scores and model sizes. With the exception of BiLSTM and ResNet18 on CSI-Bench, all achieve near-perfect classification performance, indicating only minor differences between the two dataset families.

\begin{table}[t]
    \centering
    \setlength{\tabcolsep}{3pt}
    \caption{Macro F1 (\%) on clean in-distribution test sets}
    \label{tab:clean_csi_ntu_supervised}
    \begin{tabularx}{\linewidth}{l l cccc}
        \toprule
        & & \multicolumn{2}{c}{\textbf{NTU-Fi}} & \multicolumn{2}{c}{\textbf{CSI-Bench}} \\
        \cmidrule(lr){3-4} \cmidrule(lr){5-6}
        \textbf{Model} & \textbf{Params.} & \textbf{HAR} & \textbf{HID} & \textbf{HAR} & \textbf{HID} \\
        \midrule
        MLP         & 11.2M--96.0M   & 97.0 & 97.1  & 90.0 & 100.0  \\
        LeNet       & 603.4K--3.3M   & 98.5 & 99.8  & 94.6 & 100.0  \\
        BiLSTM      & 913.0K--11.9M  & 98.1 & 99.7  & 76.5 & 99.7   \\
        ResNet18    & 11.2M          & 92.6 & 100.0 & 82.1 & 100.0  \\
        TimeSformer & 1.5M--8.1M     & 98.9 & 98.6  & 91.7 & 100.0  \\
        \bottomrule
    \end{tabularx}
\end{table}

\subsection{White-Box Robustness Benchmarking}
\label{sec:whitebox}

We next evaluated model robustness under full-knowledge (white-box) adversaries using APGD for untargeted HAR and APGD-T for targeted HID, under both unconstrained and constrained perturbation settings. Each attack was executed at SNR ``budgets" of 40~dB, 20~dB, 10~dB, and 5~dB. Table~\ref{tab:study1_whitebox_summary} summarizes the Attack Success Rates (ASRs) observed for the different configurations evaluated. 

We make the following key observations: \textit{1)} increasing the perturbation budget consistently increases ASR, with 40~dB being the most restrictive regime and 5~dB the strongest one; for example, HAR rises from $3.0 \pm 3.0$ to $64.8 \pm 36.2$ on NTU-Fi and from $13.2 \pm 25.5$ to $69.4 \pm 31.0$ on CSI-Bench; \textit{2)} HAR is substantially more vulnerable than HID across both benchmarks, so untargeted attacks are much more successful than pairwise targeted attacks in this study; \textit{3)} dataset effects are task-dependent rather than uniform: CSI-Bench is more vulnerable on HAR, whereas NTU-Fi is more vulnerable on HID at every budget; \textit{4)} applying constraints substantially suppresses HAR attack success ($53.5 \pm 44.6 \rightarrow 27.1 \pm 28.3$ on NTU-Fi and $75.4 \pm 35.3 \rightarrow 19.3 \pm 22.8$ on CSI-Bench), but the same reduction is not uniform for HID, where ASR remains low overall and CSI-HID increases slightly under constraints ($2.2 \pm 6.6 \rightarrow 5.4 \pm 11.7$); \textit{5)} model family has a strong but non-monotonic effect on robustness: the MLP is the most robust on average across all four columns, while TimeSformer/ResNet18 are the most vulnerable on HAR and LeNet is the most vulnerable on NTU-HID; \textit{6)} the large standard deviations, especially on HAR and NTU-HID, indicate substantial heterogeneity across settings, suggesting that vulnerability is driven by the interaction of model family, task, and budget rather than model size alone.

\begin{table*}[t]
\centering
\caption{Mean ($\mu$) $\pm$ SD of Adversarial Success Rate (ASR) for white-box attacks. Values are reported in percent without the percent symbol.}
\label{tab:study1_whitebox_summary}
\begin{tabular*}{\textwidth}{@{\extracolsep{\fill}}lcccc@{}}
\toprule
& \multicolumn{2}{c}{\textbf{NTU-Fi}} & \multicolumn{2}{c}{\textbf{CSI-Bench}} \\
\cmidrule(lr){2-3} \cmidrule(lr){4-5}
\textbf{Setting} & \textbf{HAR} & \textbf{HID} & \textbf{HAR} & \textbf{HID} \\
\midrule
\multicolumn{5}{@{}l}{\textbf{Effect of perturbation budget}} \\
40 dB & $3.0 \pm 3.0$ & $0.4 \pm 0.9$ & $13.2 \pm 25.5$ & $0.1 \pm 0.2$ \\
20 dB & $34.5 \pm 36.0$ & $1.0 \pm 1.3$ & $47.7 \pm 46.4$ & $0.2 \pm 0.2$ \\
10 dB & $58.9 \pm 37.5$ & $11.4 \pm 22.5$ & $59.2 \pm 38.3$ & $1.4 \pm 1.5$ \\
5 dB & $64.8 \pm 36.2$ & $17.6 \pm 31.2$ & $69.4 \pm 31.0$ & $13.4 \pm 15.8$ \\
\midrule
\multicolumn{5}{@{}l}{\textbf{Effect of model family}} \\
MLP & $4.2 \pm 2.6$ & $0.5 \pm 0.6$ & $29.0 \pm 32.2$ & $0.0 \pm 0.1$ \\
LeNet & $45.6 \pm 38.2$ & $33.2 \pm 35.4$ & $50.2 \pm 44.6$ & $5.6 \pm 14.3$ \\
BiLSTM & $33.3 \pm 36.7$ & $2.6 \pm 1.2$ & $49.4 \pm 42.7$ & $4.9 \pm 8.5$ \\
ResNet18 & $57.8 \pm 40.9$ & $1.1 \pm 1.9$ & $63.0 \pm 40.4$ & $8.0 \pm 13.1$ \\
TimeSformer & $60.6 \pm 41.1$ & $0.6 \pm 1.2$ & $45.2 \pm 46.1$ & $0.3 \pm 0.3$ \\
\midrule
\multicolumn{5}{@{}l}{\textbf{Effect of constraints}} \\
Unconstrained & $53.5 \pm 44.6$ & $8.2 \pm 20.7$ & $75.4 \pm 35.3$ & $2.2 \pm 6.6$ \\
Constrained & $27.1 \pm 28.3$ & $7.0 \pm 19.6$ & $19.3 \pm 22.8$ & $5.4 \pm 11.7$ \\
\bottomrule
\end{tabular*}
\end{table*}

\subsection{Black-Box Transfer Attacks}
\label{sec:blackbox}

\begin{table*}[t]
\centering
\caption{Mean ($\mu$) $\pm$ SD of Attack Success Rate (ASR) for black-box transfer attacks.}
\label{tab:study2_transfer_summary}
\begin{tabular*}{\textwidth}{@{\extracolsep{\fill}}lcccc@{}}
\toprule
& \multicolumn{2}{c}{\textbf{NTU-Fi}} & \multicolumn{2}{c}{\textbf{CSI-Bench}} \\
\cmidrule(lr){2-3} \cmidrule(lr){4-5}
\textbf{Setting} & \textbf{HAR} & \textbf{HID} & \textbf{HAR} & \textbf{HID} \\
\midrule
\multicolumn{5}{@{}l}{\textbf{Effect of perturbation budget}} \\
40 dB & $0.1 \pm 0.2$ & $0.0 \pm 0.0$ & $0.2 \pm 0.3$ & $0.0 \pm 0.0$ \\
20 dB & $2.1 \pm 2.2$ & $0.1 \pm 0.3$ & $3.6 \pm 4.8$ & $0.0 \pm 0.0$ \\
10 dB & $7.7 \pm 7.4$ & $1.4 \pm 2.8$ & $13.0 \pm 13.5$ & $0.0 \pm 0.0$ \\
5 dB  & $14.5 \pm 10.5$ & $1.7 \pm 3.5$ & $22.1 \pm 17.1$ & $0.1 \pm 0.2$ \\
\midrule
\multicolumn{5}{@{}l}{\textbf{Effect of model family as target}} \\
MLP         & $7.5 \pm 9.6$  & $0.0 \pm 0.0$ & $4.7 \pm 9.1$  & $0.0 \pm 0.0$ \\
LeNet       & $8.8 \pm 12.0$ & $4.1 \pm 3.8$ & $10.9 \pm 14.4$ & $0.1 \pm 0.2$ \\
BiLSTM      & $3.4 \pm 5.0$  & $0.0 \pm 0.0$ & $9.6 \pm 11.4$ & $0.0 \pm 0.0$ \\
ResNet18    & $6.1 \pm 7.7$  & $0.0 \pm 0.0$ & $15.8 \pm 17.0$ & $0.1 \pm 0.2$ \\
TimeSformer & $4.8 \pm 6.1$  & $0.0 \pm 0.0$ & $7.6 \pm 14.9$ & $0.0 \pm 0.1$ \\
\midrule
\multicolumn{5}{@{}l}{\textbf{Effect of constraints}} \\
Unconstrained & $8.8 \pm 10.3$ & $0.8 \pm 2.4$ & $14.4 \pm 16.9$ & $0.0 \pm 0.1$ \\
Constrained   & $3.5 \pm 5.2$  & $0.8 \pm 2.3$ & $5.0 \pm 8.0$  & $0.1 \pm 0.2$ \\
\bottomrule
\end{tabular*}
\end{table*}

We then turned to the black-box transfer setting, where adversarial examples are crafted on a surrogate source model and evaluated on a different target model trained on the same dataset. We report APGD-based transfer results for untargeted HAR and targeted HID across all source$\rightarrow$target model pairs, under both unconstrained and constrained settings, at SNR budgets of 40~dB, 20~dB, 10~dB, and 5~dB.

We make the following key observations from Table~\ref{tab:study2_transfer_summary}: \textit{1)} black-box transferability is substantially weaker than white-box vulnerability overall, with transfer ASR remaining near zero at 40~dB across all four columns; \textit{2)} larger perturbation budgets increase transferability, especially for HAR, where NTU-HAR rises from $0.1 \pm 0.2$ at 40~dB to $14.5 \pm 10.5$ at 5~dB, and CSI-HAR rises from $0.2 \pm 0.3$ to $22.1 \pm 17.1$; \textit{3)} HAR transfers much more readily than HID, indicating that untargeted cross-model attacks are feasible at stronger budgets, whereas targeted pairwise transfer remains difficult; \textit{4)} CSI-Bench HAR is the most transferable regime overall, while CSI-HID is effectively non-transferable under this setup, staying at approximately $0\%$ across all budgets; \textit{5)} when aggregated over source models, the most vulnerable target family depends on the dataset and task, with LeNet being the most vulnerable target on NTU-HAR and NTU-HID, and ResNet18 being the most vulnerable target on CSI-HAR; \textit{6)} applying constraints reduces transferability noticeably for HAR ($8.8 \pm 10.3 \rightarrow 3.5 \pm 5.2$ on NTU-HAR and $14.4 \pm 16.9 \rightarrow 5.0 \pm 8.0$ on CSI-HAR), while HID transfer remains very low regardless of whether it is constrained.

\subsection{Universal Adversarial Perturbations}
\label{sec:uap}

We next investigated UAPs, a stronger black-box setting in which a single perturbation is optimized to generalize across many samples rather than being tailored to each input individually. All UAP experiments were conducted at a fixed perturbation budget of 10~dB SNR, under both unconstrained and constrained settings. 

Several clear trends emerge from Table~\ref{tab:study3_uap_summary}: \textit{1)} at 10~dB, UAPs are highly effective for HAR on both benchmarks, with mean ASR of $44.7 \pm 24.7$ on NTU-HAR and $46.4 \pm 33.3$ on CSI-HAR; \textit{2)} HID is far more resistant than HAR, especially on NTU-Fi, where UAP success remains negligible ($0.4 \pm 0.7$), while CSI-HID is somewhat more vulnerable but still much lower than HAR ($5.5 \pm 9.2$); \textit{3)} realism constraints substantially reduce UAP efficacy, particularly for HAR, where ASR drops from $59.1 \pm 26.9$ to $30.3 \pm 11.2$ on NTU-Fi and from $72.3 \pm 23.8$ to $20.5 \pm 15.8$ on CSI-Bench; \textit{4)} vulnerability varies across model families, with ResNet18 and TimeSformer being the most vulnerable on NTU-HAR, and BiLSTM being the most vulnerable on CSI-HAR and CSI-HID; \textit{5)} CSI-Bench models are generally at least as vulnerable as NTU-Fi models under UAPs, and markedly more vulnerable for HID. Overall, UAPs represent a serious threat to HAR systems even with a single shared perturbation, but their effectiveness on HID is weaker and strongly dependent on both dataset and model family.

\begin{table*}[t]
\centering
\caption{Mean ($\mu$) $\pm$ SD of Attack Success Rate (ASR) for black-box Universal Adversarial Perturbation attacks.}
\label{tab:study3_uap_summary}
\begin{tabular*}{\textwidth}{@{\extracolsep{\fill}}lcccc@{}}
\toprule
& \multicolumn{2}{c}{\textbf{NTU-Fi}} & \multicolumn{2}{c}{\textbf{CSI-Bench}} \\
\cmidrule(lr){2-3} \cmidrule(lr){4-5}
\textbf{Setting} & \textbf{HAR} & \textbf{HID} & \textbf{HAR} & \textbf{HID} \\
\midrule
\multicolumn{5}{@{}l}{\textbf{Effect of perturbation budget}} \\
10 dB & $44.7 \pm 24.7$ & $0.4 \pm 0.7$ & $46.4 \pm 33.3$ & $5.5 \pm 9.2$ \\
\midrule
\multicolumn{5}{@{}l}{\textbf{Effect of model family}} \\
MLP         & $34.3 \pm 12.6$ & $0.5 \pm 0.3$ & $46.4 \pm 59.7$ & $3.0 \pm 4.3$ \\
LeNet       & $26.6 \pm 11.6$ & $0.0 \pm 0.0$ & $38.6 \pm 11.3$ & $0.8 \pm 1.1$ \\
BiLSTM      & $46.8 \pm 39.0$ & $0.0 \pm 0.0$ & $65.3 \pm 35.3$ & $13.5 \pm 18.8$ \\
ResNet18    & $57.6 \pm 27.2$ & $0.2 \pm 0.2$ & $56.0 \pm 48.2$ & $1.4 \pm 1.8$ \\
TimeSformer & $58.1 \pm 36.7$ & $1.1 \pm 1.5$ & $25.6 \pm 28.6$ & $8.7 \pm 12.2$ \\
\midrule
\multicolumn{5}{@{}l}{\textbf{Effect of constraints}} \\
Unconstrained & $59.1 \pm 26.9$ & $0.6 \pm 0.9$ & $72.3 \pm 23.8$ & $10.9 \pm 10.8$ \\
Constrained   & $30.3 \pm 11.2$ & $0.1 \pm 0.2$ & $20.5 \pm 15.8$ & $0.1 \pm 0.1$ \\
\bottomrule
\end{tabular*}
\end{table*}

\subsection{Adversarial Defenses}
\label{sec:advtrain}

\begin{table*}[t]
\centering
\caption{Mean ($\mu$) of Clean Accuracy, Empirical Robust Accuracy, and Certified Performance for defenses.}
\label{tab:study4_defence_summary_combined}
\begin{tabular*}{\textwidth}{@{\extracolsep{\fill}}lcccc@{}}
\toprule
& \multicolumn{2}{c}{\textbf{NTU-Fi}} & \multicolumn{2}{c}{\textbf{CSI-Bench}} \\
\cmidrule(lr){2-3} \cmidrule(lr){4-5}
\textbf{Metric} & \textbf{HAR} & \textbf{HID} & \textbf{HAR} & \textbf{HID} \\
\midrule
\multicolumn{5}{@{}l}{\textbf{APGD-Adversarial Training}} \\
Clean Acc.  & $95.2 \pm 4.8$ & $95.8 \pm 3.6$ & $84.4 \pm 1.5$ & $97.6 \pm 5.0$ \\
Robust Acc. & $36.3 \pm 41.0$ & $99.9 \pm 0.2$ & $9.0 \pm 12.6$ & $98.5 \pm 3.3$ \\
\midrule
\multicolumn{5}{@{}l}{\textbf{Randomized Smoothing}} \\
Clean Acc.  & $93.7 \pm 4.4$ & $96.5 \pm 1.5$ & $77.8 \pm 15.3$ & $99.6 \pm 0.7$ \\
Robust Acc. & $28.6 \pm 40.5$ & $97.8 \pm 4.0$ & $22.5 \pm 30.6$ & $99.8 \pm 0.5$ \\
Cert. Acc.  & $82.1 \pm 14.8$ & $82.5 \pm 19.9$ & $39.1 \pm 16.3$ & $86.1 \pm 30.9$ \\
Mean Radius & $0.59 \pm 0.00$ & $0.59 \pm 0.00$ & $0.49 \pm 0.10$ & $0.59 \pm 0.01$ \\
\bottomrule
\end{tabular*}
\end{table*}

We evaluated two robustness-augmentation strategies against a strong 10~dB APGD-$\ell_2$ threat (200 steps, 10 restarts). Table~\ref{tab:study4_defence_summary_combined} summarizes the resulting clean, robust, and certified performance across the five model families, allowing us to assess the trade-offs introduced by each defense. Both defenses retain strong clean accuracy overall, but their effectiveness is fundamentally task-dependent. AT is highly effective for HID, where robust accuracy is nearly saturated across both benchmarks, but it is much less successful for HAR, particularly on CSI-Bench, where robust performance drops sharply. RS, by contrast, emerges as the more comprehensive defense: it matches the strong HID robustness of AT while additionally providing formal certification, and it offers a more favorable clean--robust--certified trade-off for the much more vulnerable HAR setting. Overall, these results reinforce that HID is intrinsically easier to defend than HAR, and that RS provides the strongest overall security posture.

\subsection{Analysis of Adversarial Samples}

\begin{figure*}[t]
    \centering
    \includegraphics[width=0.81\linewidth]{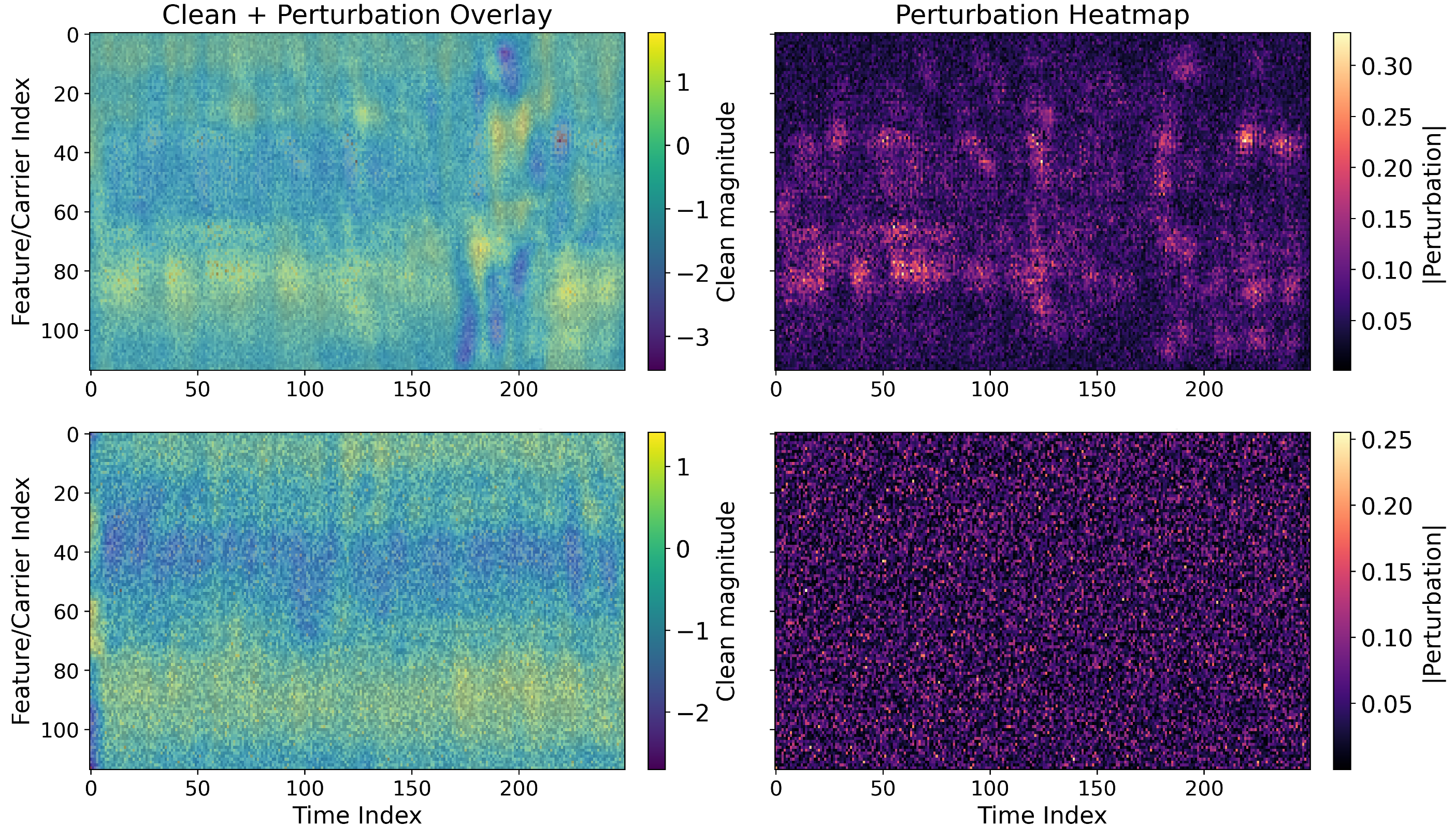}
    \caption{NTU-Fi-HAR: \emph{walk} $\rightarrow$ \emph{box} without (\textit{above}) and with (\textit{below}) constraints: $\mathrm{\ell_2}$ attack ($\mathrm{SNR}=20$\,dB)}
    \label{fig:walk_untargeted_heatmap}
\end{figure*}

Figure~\ref{fig:walk_untargeted_heatmap} contrasts unconstrained and physically constrained APGD-$\ell_2$ perturbations, where each pixel denotes the mean absolute perturbation $|\boldsymbol{\delta}|$ across time (horizontal) and subcarriers (vertical). In the unconstrained setting (top), the optimizer exploits mathematical loopholes to surgically target specific, physically meaningful subcarrier groups and temporal windows, with impossible precision. Conversely, enforcing spectral-temporal and cross-antenna constraints (bottom) prevents the independent manipulation of isolated features. Consequently, the perturbation energy must be masked by the natural stochasticity of the wireless channel.

\section{Discussion}
\subsection{Design Principles for Robust Sensing}

Our empirical findings lead to several actionable principles:

\begin{itemize}
    \item \textbf{Robustness does not increase monotonically with model size:} In our evaluated model families, simpler baselines such as the MLP often outperform deeper architectures such as ResNet18 and TimeSformer under adversarial perturbation.

    \item \textbf{Robustness depends strongly on task type:} Across both benchmark families, HAR models are substantially more vulnerable than HID models under white-box, transfer-based, and universal attacks.

    \item \textbf{Constrained attacks yield more conservative robustness estimates:} Physics-inspired temporal, spectral, and cross-antenna constraints substantially reduce ASR relative to unconstrained perturbations.

    \item \textbf{Universal perturbations are important for HAR:} In the evaluated digital setting, UAPs severely degrade untargeted HAR performance, although their effectiveness drops substantially with constraints.

    \item \textbf{Robust training helps but is incomplete:} The gains---clean marginally down, robust up---remain strongly task- and model-dependent; no defense consistently eliminates the clean–robustness trade-off.
\end{itemize}

\subsection{Hardware and Deployment Constraints}

Beyond accuracy, practical deployment on edge platforms is limited by compute, memory, and energy. Our benchmarking experiments revealed that there are three critical operational trade-offs:

\begin{itemize}
    \item \textbf{Storage cost and compute cost diverge sharply:} Parameter-efficient models are not necessarily compute-efficient on high-dimensional CSI inputs. While the MLP carries a massive weight footprint (e.g., 42.76~MiB in FP32 for NTU-HAR) compared to LeNet (12.43~MiB), its estimated edge inference latency is dramatically lower (1.12~ms vs.\ 446.31~ms). This pattern repeats on CSI-HAR, where the MLP requires only 1.85~ms, while smaller models like LeNet and BiLSTM bottleneck at 84.34~ms and 76.41~ms. Deployments must strictly weigh memory capacity against latency.

    \item \textbf{The necessity (and risk) of quantization:} Because fast, low-latency architectures like the MLP often exceed embedded memory limits, INT8 post-training quantization is practically mandatory ($4$x memory reduction). However, this compression risks corrupting the fragile amplitude--phase structures of CSI data, degrading both accuracy and robustness.

    \item \textbf{Physical constraints heavily tax the attacker:} Enforcing physical stealth acts as a two-fold defense---it severely degrades attack efficacy while forcing the adversary to spend significantly more compute. For example, applying constrained APGD against BiLSTM on CSI-HAR increases per-sample generation time from 11.5~ms to 47.3~ms, while attack success (ASR) plummets from 98.4\% to 30.3\%. Similarly, constrained attacks against LeNet on NTU-HAR raise generation time from 103.9~ms to 279.3~ms, while ASR falls from 96.9\% to 51.2\%.
\end{itemize}

Overall, no single architecture currently offers a perfect deployment profile for wireless sensing. Models with the smallest memory footprints (LeNet, BiLSTM) suffer from severe latency bottlenecks, whereas the fastest models (MLP) demand prohibitive memory. Future work must prioritize hardware-aware co-design to identify architectures that balance latency, memory, and resistance to constrained attacks.

\subsection{Problem-Space Adversarial Attacks}

The reduction in attack success under physics-inspired constraints highlights an important distinction between unconstrained digital perturbations and attacks that respect wireless signal structure. In wireless sensing, propagation and receiver processing impose temporal, spectral, and spatial regularities that arbitrary \emph{feature-space} perturbations ignore. This suggests that unconstrained digital attacks may overstate practical vulnerability.

At the same time, our constrained attacks remain a proxy rather than a full \emph{problem-space}~\cite{pierazzi_intriguing_2020} implementation. We do not optimize end-to-end OTA perturbations, in part because the wireless channel, hardware pipeline, and CSI extraction process introduce non-differentiable components that complicate gradient-based optimization. Instead, our approach estimates structured digital perturbations that better approximate feasible signal distortions after wireless propagation. Bridging this gap---potentially through standardized differentiable approximations of the physical channel---constitutes a critical direction for future research.

\subsection{Implications of Feature Engineering}

CSI-Bench~\cite{zhu_csi-bench_2025} shows that models trained on raw CSI perform substantially worse under cross-device, cross-user, and cross-environment evaluation. In parallel, pipelines based on feature engineering such as Widar3.0~\cite{zheng_zero-effort_2019} and SHARP~\cite{meneghello_sharp_2023} aim to derive more domain-invariant representations that improve generalization. An open question is how such feature engineering interacts with adversarial robustness. On one hand, engineered features may suppress unwanted variation and improve cross-domain stability. On the other hand, they may remove low-level cues that contribute to robust class separation under adversarial perturbation. Future work should therefore compare robustness across raw and engineered CSI pipelines, and examine whether generalization and adversarial robustness can be jointly improved through robust feature alignment or invariance-driven learning.

\subsection{Attacks in the Time Domain}

Many public CSI datasets rely on assumptions such as indoor-only capture, fixed transceiver pairs, limited scene diversity, and perfectly segmented activities. While these assumptions improve reproducibility, they leave open important robustness questions for continuous sensing deployments. In dynamic real-world environments, attackers may exploit temporal correlation and online adaptation effects rather than relying only on isolated test-time perturbations. This motivates extending future robustness evaluations toward temporally persistent attacks, adaptive drift injection, and online adversarial learning scenarios that better reflect continuous Wi-Fi sensing systems.

\subsection{Utility of a Modular Framework}

A key contribution of this work is the establishment of a consistent, modular, and extensible benchmarking framework for adversarial robustness in wireless sensing. In the same spirit as benchmark-driven progress in other machine learning domains, such a framework provides a reproducible basis for iterative improvement through the cycle of \emph{attack $\rightarrow$ defense $\rightarrow$ re-evaluation}. Our open-source implementation supports the integration of new CSI datasets, architectures, and defenses under a common evaluation pipeline. This foundation enables systematic study of open questions including cross-dataset transferability, multi-modal robustness, and efficient defenses for edge deployments, and may help move the field toward more trustworthy ML systems.

\subsection{Limitations}

This work has two key limitations. First, our experiments focus on attacks applied after CSI preprocessing rather than implementing end-to-end OTA attacks on the physical sensing pipeline; validating the proposed physics-inspired projection operator against real wireless attack data like \cite{zhou_wiadv_2022, li_practical_2024} therefore remains an important next step. Second, our evaluation targets raw CSI datasets and does not yet extend to engineered representations such as BVP (Widar3.0 \cite{zheng_zero-effort_2019}) or Doppler-based features (SHARP \cite{meneghello_sharp_2023}), which may potentially exhibit different robustness profiles.

\section{Conclusions}
As CSI sensing becomes central to future Wi-Fi standards (e.g., IEEE 802.11bf), its ML layer is growing as a critical attack surface. To systematically evaluate this, we introduced a unified framework benchmarking adversarial robustness across diverse models, datasets, attacks and defenses. Our evaluation yields four key takeaways. First, robustness does not scale monotonically with model size. Second, adversarial vulnerability is highly task-dependent, with Human Activity Recognition systems proving consistently more susceptible than Human Identification systems. Third, standard unconstrained attacks drastically overestimate real-world vulnerability. Enforcing temporal, spectral, and spatial constraints sharply reduces attack success, providing a much more realistic stress test for Over-The-Air (OTA) realizability. Finally, practical deployment introduces a severe hardware trade-off: no single evaluated architecture currently balances clean accuracy, adversarial robustness, and edge constraints (latency, memory, and quantization). Delivering trustworthy CSI sensing requires shifting focus from standard accuracy to hardware-aware co-design and physically constrained robustness evaluation. We plan to open-source our modular benchmarking framework to accelerate community-driven research toward secure, deployable wireless sensing systems.

\section*{Acknowledgments}
Shreevanth Krishnaa Gopalakrishnan was supported by the EPSRC through the Centre for Doctoral Training Studentship in Cybersecurity (EP/S022503/1).

\printbibliography

\end{document}